\documentclass{Interspeech2024}
\usepackage{multirow}
\usepackage{placeins}
\usepackage{amsmath}

\interspeechcameraready

\title{
 Connected Speech-Based Cognitive Assessment in Chinese and English}

\name[affiliation={1}]{Saturnino}{Luz}
\name[affiliation={2}]{Sofia}{De La Fuente Garcia}
\name[affiliation={3}]{Fasih}{Haider}
\name[affiliation={4}]{Davida}{Fromm}
\name[affiliation={4}]{Brian}{MacWhinney}
\name[affiliation={5}]{Alyssa}{Lanzi}
\name[affiliation={6}]{Ya-Ning}{Chang}
\name[affiliation={7}]{Chia-Ju}{Chou}
\name[affiliation={7}]{Yi-Chien}{Liu}

\address{\small
  $^1$Usher Institute,  Medical School,  
  $^2$School of Health in Social Science,
  $^3$School of Engineering, University of Edinburgh, UK\\
  $^4$Department of Psychology, Carnegie Mellon University,  
  $^5$Communication Sciences \& Disorders, University of Delaware, USA \\
  $^6$Miin Wu School of Computing  National Cheng Kung University, 
  $^7$Department of Neurology, Cardinal Tien Hospital, Taipei, Taiwan\vspace*{-.5ex}}
\email{s.luz@.ed.ac.uk}


\usepackage[textsize=tiny,disable]{todonotes}
\usepackage{soul}
\newcommand{\Mnewauthor}[3][]{%
    \def\@tempa{#1}%
    \ifx\@tempa\@empty%
        \def\@authid{#2}%
    \else%
        \def\@authid{#1}%
    \fi%
    \expandafter\newcommand\csname mn\@authid\endcsname[2][]{%
      \if@mnotes@hide ##1\else\sethlcolor{#3}\hl{##1}\todo[color=#3]{#2: ##2}{}\fi}%
}
\Mnewauthor[SL]{SL}{yellow}
\Mnewauthor[SG]{SG}{blue}
\Mnewauthor[FH]{FH}{orange}
\newcommand{\red}[1]{\textcolor{red}{#1}}

\begin{document}

\maketitle

\begin{abstract}
  We present a novel benchmark dataset and prediction tasks for
  investigating approaches to assess cognitive function through
  analysis of connected speech. The dataset consists of speech samples
  and clinical information for speakers of Mandarin Chinese and
  English with different levels of cognitive impairment as well as
  individuals with normal cognition. These data have been carefully
  matched by age and sex by propensity score analysis to ensure
  balance and representativity in model training. The prediction tasks
  encompass mild cognitive impairment diagnosis and cognitive test
  score prediction. This framework was designed to encourage the
  development of approaches to speech-based cognitive assessment which
  generalise across languages. We illustrate it by presenting baseline
  prediction models that employ language-agnostic and comparable
  features for diagnosis and cognitive test score
  prediction. Unweighted average recall was 59.2\% in diagnosis, and
  root mean squared error was 2.89 in score prediction.
\end{abstract}

\section{Introduction}
Cognitive problems such as memory loss, speech and language
impairment, and reasoning difficulties occur frequently among older
adults and often precede the onset of dementia syndromes. Due to the
high prevalence of dementia and the costs this implies to
health systems worldwide \cite{Nichols2022Feb},
research into cognitive impairment for the purposes of dementia
prevention and early detection has become a priority in
healthcare. There is a need for cost-effective and scalable methods
for assessment of cognition and detection of impairment, from its most
mild forms to severe manifestations of dementia. Speech is an easily
collectable behavioural signal which reflects cognitive function and,
therefore, could potentially serve as a digital biomarker of cognitive
function, presenting a unique opportunity for application of speech
technology \cite{bib:DelaFuenteRichieLuz2020JAD}.

We aim to assess speech as a behavioural marker of cognition in a
global health context by investigating its application to the
modelling of cognitive health indicators in two major languages,
namely, Chinese and English. In this paper, we focus on prediction of
cognitive test scores and diagnosis of mild cognitive impairment (MCI)
in older speakers of Chinese and English, using samples of connected
speech produced in picture description tasks. Our aim was to investigate
approaches that are language independent or build on comparable
features. To this end, we created, and are sharing with the research
community, recorded speech from study participants doing picture
description tasks along with clinical and neuropsychological test
data.

This dataset has been used as a benchmark for speech processing and
machine learning tasks that are relevant to the detection of cognitive
decline through analysis of connected speech data. It formed the basis
of the TAUKADIAL Challenge, at Interspeech 2024
(http://luzs.gitlab.io/taukadial/). We hope that this new resource
will stimulate research on speech biomarkers in the speech, signal
processing, machine learning and
biomedical research communities, enabling them to test existing
methods or develop novel approaches on a new, standardised dataset
which will remain available to the community for future research and
replication of results.

\section{Background}

The field of speech-based approaches to detecting cognitive decline
has grown considerably over the last two decades, with a major focus
on detecting dementia or Alzheimer's dementia (AD) in comparison to a
control (neurotypical or normal cognition, NC) group. A smaller
proportion of studies has focused on MCI detection
\cite{bib:DelaFuenteRichieLuz2020JAD}. Most studies report accuracy
figures without class balance, where accuracy is a biased measure. For
example, \cite{bib:Nasrolahzadeh2018} report a relatively high
accuracy, 97.71\%, in a highly imbalanced dataset while
\cite{bib:Mirheidari2019computational} and \cite{bib:Mirzaei2018}
report comparably lower accuracy, 62\%, in a more balanced datasets.
Similarly, using speech data generated from a cognitive assessment
(picture description task), \cite{bib:Guo2019} obtained 85.4\%
accuracy using text-based features only on an imbalanced set of 268
participants. In contrast, \cite{bib:HaiderFuenteLuz20aspacf}
generated a subset of the same data (164 participants), balanced for
class, gender, and age, and reported 78.7\% accuracy with only acoustic
features from standardised feature sets developed for computational
paralinguistics.

Clinical tests such as the Mini-Mental State Examination (MMSE) are
often part of these studies as mere data descriptors, rarely used in
prediction. Some studies
\cite{bib:Prudhommeaux2015,bib:Sadeghian2017,bib:Shinkawa2019,jin2023consen}
have used MMSE results as a baseline for classification, against which
to compare the speech-based classifier, but very few available studies
go beyond classification and use speech-based approaches to predict
MMSE scores or other cognitive tests. However, there has been a shift
of focus toward it in recent years. For instance,
lexico-semantic features extracted from picture descriptions have been
used in a model that was able to explain 51\% of the variance of
cognitive scores at the time of speech collection, and 56\% of
cognitive scores in a 12-month follow-up \cite{bib:ostrand2021using}.
Other approaches have also been published, such as
\cite{bib:liu2021automatic}, which used BERT to predict MMSE scores
from denoised speech recordings from picture description tasks and
reported a root mean squared error (RMSE) of 3.76. Another study
reported that acoustic features alone predict MMSE scores with a mean
absolute error (MAE) of 5.66 and an $R^2$ of 0.125, with a linear
regression analysis that improved by adding age, sex, and years of
education to the model, yielding a MAE = 4.97 and $R^2$ = 0.261
\cite{bib:fu2020predicting} on the balanced dataset used by
\cite{bib:HaiderFuenteLuz20aspacf}.




None of these studies addresses multilingual models where research is
scarce and heterogeneous. A study on the AZTIAHO database reported
accuracy scores between 60\% and 93.79\% using only \textit{ad hoc}
acoustic features. While this database contains samples in English,
French, Spanish, Catalan, Basque, Chinese, Arabian, and Portuguese, it
is small (40 participants) and remarkably age- and class-imbalanced
\cite{bib:Lopez-de-Ipina2015b}.

Another multilingual study used English and Swedish speech samples
generated through picture description tasks and word embeddings to
train models that obtained classification accuracy of 63\% for English
and 72\% for Swedish \cite{bib:fraser2019multilingualMCI} in MCI
diagnosis, and 75\% ($F_1=0.77$) in AD AD/NC classification on 57
French and 550 English samples \cite{bib:fraser2019multilingualAD}.
More recently, a signal processing grand challenge addressed the issue
of generalising speech-based predictive models across two languages:
Greek and English \cite{bib:LuzHaiderEtAl23madress}. Differently from
our experimental setting, theirs involved training of models in one
language and testing on another. The top performing systems had
classification accuracy between 69\% to 87\% (AD vs NC) and MMSE
score prediction errors RMSE between 4.79 and 3.72. To the best of our
knowledge, this study addresses a gap in the literature by combining
multilingual speech analysis, MCI detection, and prediction of
cognitive scores.



\section{Data}

Speech data are most often obtained from tasks embedded in
neuropsychological batteries.  Our dataset consists of Chinese and
English speech samples collected while the speakers participated in
picture description tasks conducted as part of cognitive assessments
in clinical settings.

English-speaking participants were recruited from a community in the
United States through print and online advertisements targeted to
adults aged 60-90 with memory concerns. Eligible participants were at
least 60 years old, spoke and understood English, had adequate hearing
and vision to participate in a telehealth session, were stable on or
not taking nootropic medications, and had a negative self-reported
history of major psychiatric disorder or other medical disorder or
illness that could cause cognitive decline (e.g., traumatic brain
injury). Participants were classified as either NC or MCI. To be
classified as MCI, a neuropsychologist determined that participants
met the following National Institute on Aging-Alzheimer's Association
(NIA-AA) criteria ~\cite{albert2011diagnosis}: (a) self-reported a
decline in cognition, (b) documented impairment in memory (produced a
score greater than or equal to -1.5 SD on an objective measure), c)
preserved functional independence (obtained a global score of less
than or equal to 0.5 on the Clinical Dementia Rating Scale
\cite{morris1993clinical} - interview with a loved one), and (d) not
demented. The University of Delaware Institutional Review Board
approved data collection.

After providing informed consent, participants completed an assessment
session via videoconferencing that lasted approximately 90
minutes. During this session, participants completed the discourse
protocol and cognitive-linguistic battery with an assessor
\cite{bib:LanziSaylorEtAl23d}. The discourse protocol tasks relevant
to this project are: 1) the "Cookie Theft" picture description task
\cite{kaplan1983boston} elicited with the prompt, "Please tell me
everything you see going on in this picture"; 2) the "Cat Rescue"
picture \cite{nicholas1993system} elicited with the prompt, "Tell me a
story with a beginning, a middle, and an end"; and 3) the Norman
Rockwell print "Coming and Going" \cite{rockwell1947going} elicited
with the same prompt as the Cat Rescue task. The cognitive-linguistic
battery included the MoCA \cite{nasreddine2005montreal}, whose scores
were mapped to MMSE in this dataset following accepted practice
\cite{bib:FasnachtWueestEtAl23cm}. The assessor used a standardised
script and materials to deliver the discourse protocol and
audio-recorded the administration using high-quality audio recording
guidelines. The study data collection was managed using Research
Electronic Data Capture \cite{harris2019redcap} tools.

In the study used for collecting Chinese-language data, inclusion
criteria were participants between 60 and 90 years old, with at least
six years of education, and no history of neurological or psychiatric
disorders. The neurologist evaluated participants with MCI according
to the NIA-AA criteria. The evaluation was based on their CDR scores,
which had a global score of 0.5, and brain magnetic resonance imaging
(MRI) conducted within two years before recruitment, which showed
atrophy in regions related to Alzheimer's disease.
Picture description tasks were employed to elicit connected speech, and
responses were recorded the responses using a digital recorder. Participants
described a set of three pictures depicting Taiwanese culture, with
the instruction to report everything they observed in each one. The
evaluators refrained from providing feedback but encouraged
participants to elaborate if their responses were insufficient.
Ethical approval was obtained from the Institutional Review Board of
\ifinterspeechfinal Cardinal Tien Hospital in Taipei, Taiwan
(CTH-110-3-8-041), \else [REMOVED FOR DOUBLE-BLIND REVIEW] Hospital,
\fi and all participants signed a written informed consent document.
\mnSL{Anonymise this for submission}

The full dataset (English and Chinese) was age- and gender-balanced to
avoid bias in modelling. We ensured that the speech recordings met
suitable audio quality standards for processing. Propensity score
matching \cite{bib:RosenbaumRubin83} was employed to generate an
unbiased training set. The dataset was matched to scores defined in
terms of the probability of an instance being treated as AD given
covariates age and sex estimated through logistic regression, and
matching instances were selected. All standardised mean differences
for the covariates, standardised mean differences for squares, and
two-way interactions between covariates were well below 0.1,
indicating that the resulting set was adequately balanced.

The training set contained both Chinese and English samples with three
picture descriptions per participant. The test set comprised
recordings from different participants, with the same mix of languages
and picture descriptions. Basic descriptive statistics of training and
test set are shown in Table~\ref{tab:dsdesc}. Overall, there are 507
speech samples (261 Chinese and 246 English) with total duration of
528 minutes, ratio of training to test samples is just over 3:1. The
dataset has been made available to the wider research community via
DementiaBank \cite{bib:LanziSaylorEtAl23d}.

\begin{table}[htb]
  \centering
  \caption{Dataset description, age, MMSE and duration expressed and
    mean years, score and seconds respectively. Numbers in brackets
    correspond to standard deviation and range.}\vspace*{-1ex}
  \label{tab:dsdesc}
  \begin{tabular}{@{}l@{~~}r@{~~}r@{}}
    \hline
    \multicolumn{2}{c}{MCI} & NC                                                    \\
    \hline
Age      & 73.36 (6.14, 61-87)       & 71.85 (6.65, 61-87)       \\          
Men      & 39.2\% ($n = 87$)           & 38.2\% ($n=63$)           \\           
Women    & 60.8\% ($n = 135$)          & 61.8\% ($n=102$)          \\           
MMSE     & 25.84 (3.73, 13-30)       & 29.07 (1.08, 25-30)       \\                    
Duration & 58.92 (36.6, 12.7-240.9) & 63.07 (33.9, 10.2-209.6) \\                    
        \hline
  \end{tabular}
\end{table}

\begin{figure*}[!h]
  \centering
  \includegraphics[width=.87\linewidth]{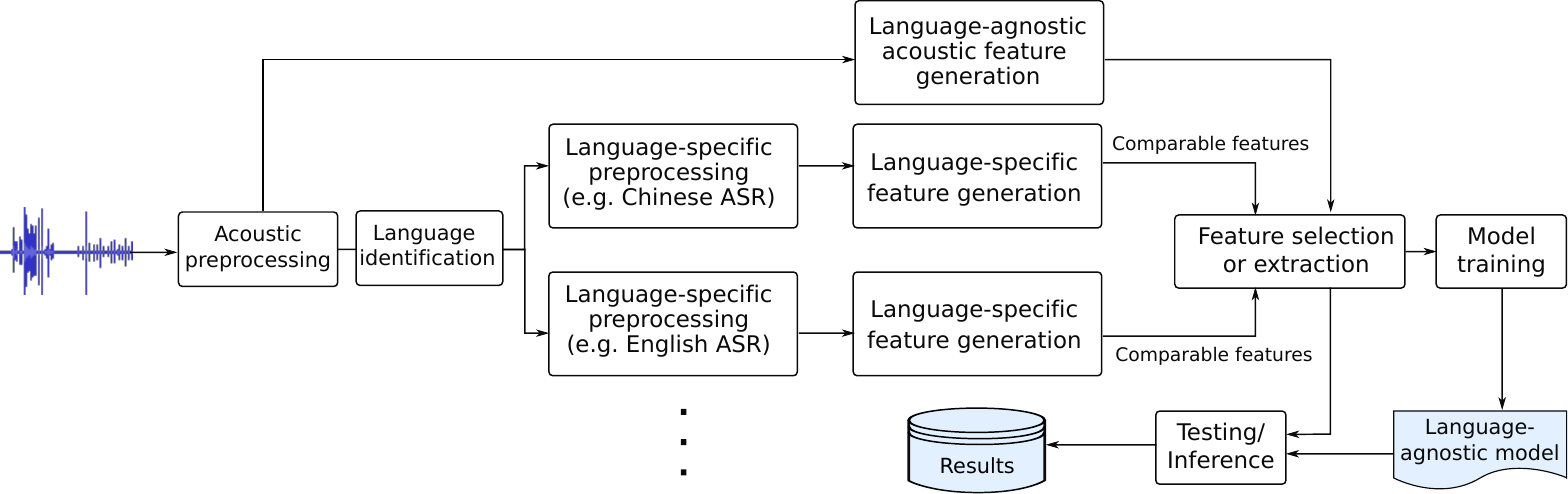}
  \caption{General architecture for multilingual cognitive assessment
    based on recorded speech.  }
  \label{fig:architecture}
\end{figure*}
\vspace{-2mm}

\section{Cognitive assessment tasks}

The benchmark presented in this paper encompasses the following tasks:
(a) a classification task, where we aimed to create models to
distinguish NC speech from MCI speech, and (b) a cognitive test score
prediction (regression) task, where we created models to infer the
subject's MMSE scores based on connected (spontaneous) speech data.

The MCI classification task is evaluated through specificity
(\(\sigma\)), sensitivity (\(\rho\)) and \(F_1\) scores for the MCI
category. These metrics are computed as follows:
$\sigma = \frac{ T_N }{T_N + F_P}$,
$F_1 = \frac{ 2 \pi \rho}{\pi + \rho}$, where
$\pi = \frac{ T_P }{T_P + F_P}$, $\rho = \frac { T_P }{T_P + F_N}$,
$N$ is the number of patients, $T_P$ is the number of true positives,
$T_N$ is the number of true negatives, $F_P$ is the number of false
positives and $F_N$ the number of false negatives. The balanced
accuracy metric (unweighted average recall, UAR) is used for the
overall ranking of this task's results. It is defined as follows:
$\operatorname {UAR} = \frac { \sigma + \rho }{2}$.

The MMSE regression task is assessed using the RMSE, defined as
$\operatorname {RMSE} =\sqrt \frac {\sum _{i=1}^{N}(\hat
  {y}_{i}-y_{i})^{2}}{N}$, where \(\hat{y}\) is the predicted MMSE
score, \(y\) is the patient's actual MMSE score, and \(\bar{y}\) is
the mean score.

\section{Modelling approach}

As our goal is to explore models that generalise across languages, we
aimed to create a single predictive model for each task which
encompassed features extracted from both languages. Thus, the general
architectures of our classification and regression systems are shown
in Figure~\ref{fig:architecture}, where {\em comparable} features
extracted from both languages are combined into a single predictive
model.

\subsection{Feature extraction}

The feature extraction procedure aimed to identify speech features
that could generalise well across the two languages. For acoustic
features, we tested two different approaches: a traditional feature
engineering approach with a feature set that has been found useful in
emotion recognition and other computational paralinguistics tasks
(eGeMAPs), and a self-supervised feature learning approach.

The eGeMAPs feature set comprises the F0 semitone, loudness, spectral
flux, MFCC, jitter, shimmer, F1, F2, F3, alpha ratio, Hammarberg
index, and slope V0 features, along with numerous statistical
functions applied to these features. This results in a total of 88
features for every audio recording \cite{bib:EybenSchererEtAl16gg}.

For self-supervised feature extraction we used the pre-trained model
wav2vec, without fine tuning, and extracted features directly from raw audio
\cite{schneider2019wav2vec}.

To balance the duration of all audio
recordings, we zero-padded the audio recordings for feature
extraction. The features are extracted from the feature extractor
layer. Then we applied a dropout layer, followed by a feature
aggregation layer and another dropout layer. For dimensionality
reduction, we used MaxPool1d layer (with a size of 42000, and a stride
of 10,000). The result was used as input features for the multilayer
perceptron (MLP) models. This resulted in 512 features per audio
recording. 

Finally, we extracted linguistic features that could be compared
across languages. The recordings were first transcribed using
automatic speech recognition (ASR) and part-of-speech tagged. Then the
following features were calculated: number of tokens, number of types,
type-to-token ratio, density (the ratio verbs, adjectives, adverbs,
prepositions, and conjunctions to the total number of tokens), verb
ratio, and pronoun ratio. To account for variability in pictures and
descriptions, the number of tokens and number of types were z-score
normalised.  \mnSL{We should also try adding comparable language
  features to these models. Perhaps this will improve the
  classification baselines. The regression baseline seems quite strong
  with wav2vec alone though.}

\subsection{Classification and regression}

\mnSL{We need to include (IS'24 checklist): ``a) An explanation of
  evaluation metrics used, b) An explanation of how models are
  initialized, if applicable.  c) Some measure of statistical
  significance of the reported gains or confidence intervals.  d) A
  description of the computing infrastructure used and the average
  runtime for each model or algorithm (e.g. training, inference etc).
  e) The number of parameters in each model.}
Multi-layer of Perceptron (MLP) models were trained on different
combinations of the above described feature sets using the Adam solver
with relu activation.  MLP models were employed for both
classification and regression.  We set $\alpha=10^{-4}$, hidden layers
of sizes 55, 160, 160 and 55, constant learning rate 0f 0.001 and a
maximum of 10,000 iterations.  In both cases, 20-fold cross-validation
was employed. The models were developed on an Intel Core
i9-9980HK CPU @ 2.40GHz 2.40 GHz with 16 GB RAM and 8 GB GPU memory
(Nvidia GeForce RTX 2080 with max-q design).
The software used for balancing the dataset, feature extraction, model
training, cross-validation and testing is available at
\ifinterspeechfinal https://gitlab.com/luzs/taukadial. \else
\url{https://gitlab.com/[REMOVED FOR DOUBLE-BLIND REVIEW]}. \fi
\mnSL{We need to add (as per Interspeech instructions: ``A description
  of the computing infrastructure used and the average runtime for
  each model or algorithm (e.g. training, inference etc), and the
  number of parameters in each model.}
\mnSL{Was hyperparameter search conducted. If so, include: f) Final
  results on a held-out evaluation set not used for hyperparameter
  tuning.  g) Hyperparameter configurations for best-performing
  models.  h) The method for choosing hyperparameter values to
  explore, and the criterion used to select among them.}
\mnSL{Upload the source code to the gitlab repo and add a note here
  incidcating how the code can be obtained and the results reproduced.}

\section{Results}

\begin{figure}
  \centering
  \includegraphics[width=.8\linewidth]{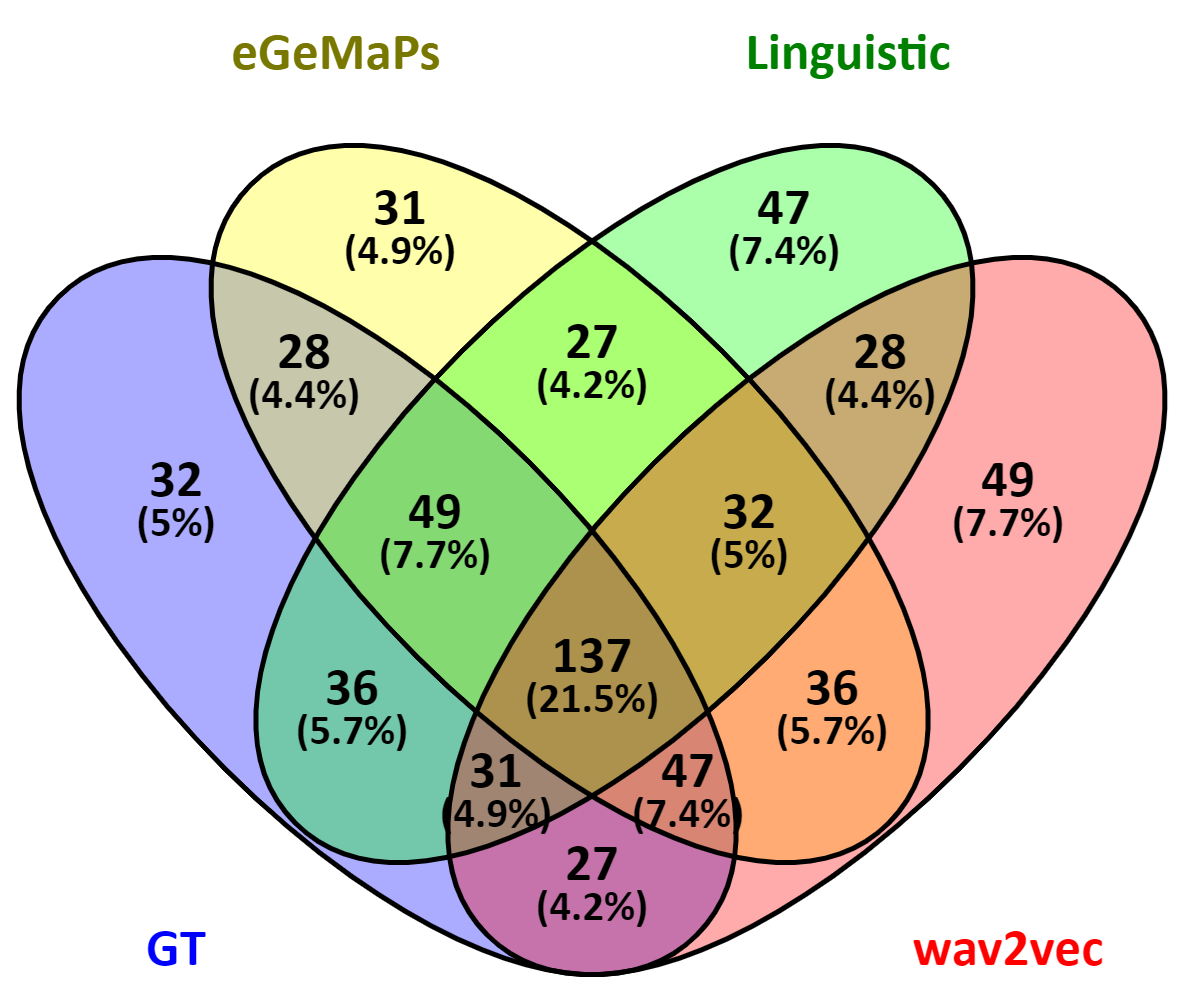}
  \caption{Venn diagram showing the effect of each features set on classification with respect to Ground Truth (GT).}
  \label{fig:venn}
\end{figure}

For the classification (diagnostic) task, our model achieved a
test-data UAR of 59.18\% while fusing the wav2vec and eGeMAPs
features. The full set of results is shown in Table
\ref{tab:results}. Confidence interval were obtained through
bootstrapping over 1000 runs \cite{bib:FerrerEieraCI24}.  The baseline
result for this task is 59.18\% UAR obtained on test data
($\sigma=0.587$, $\rho=0.597$, $\pi= 0.617$).  The results were very
similar in both languages (English: UAR$ = 60.00\%$, $\sigma=0.40$,
$\rho=0.80$; Chinese: UAR$ = 60.04\%$, $\sigma=0.39$, $\rho=0.81$).
The overall accuracy score was 0.592, while F1 reached 0.602. 
Figure \ref{fig:venn} shows
the effect of each feature set on classification performance and Table~\ref{lan_results} shows the results for both languages for the best performing methods.
\mnSL{The problem with these results is that
  if we were to choose a model purely based on CV we would have chosen
  the wav2vec, which clearly overfits, giving a test set UAR of just
  over 46\%. We need to address this.}

\begin{table*}[htb]\centering
  \caption{Summary of results for the classification task (T.1), in \% UAR,
    and the MMSE regression task (T.2), in RMSE for different
    features set combinations, where w2v = wav2vec and ling =
    comparable linguistic features, with confidence intervals in
    square brackets. }
  \label{tab:results}\vspace*{-1ex}
  \resizebox{\linewidth}{!}{
\begin{tabular}{@{}r@{~~}r@{~~}r@{~~}r@{~~}r@{~~}r@{~~}r@{~~}r@{~~}r@{}}
  \hline
  &
  & \multicolumn{1}{c}{eGeMAPs}
  & \multicolumn{1}{c}{w2v}
  & \multicolumn{1}{c}{w2v+eGeMAPs}
  & \multicolumn{1}{c}{linguistic}
  & \multicolumn{1}{c}{w2v+linguistic}
  & \multicolumn{1}{c}{ling+eGeMAPs}
  & \multicolumn{1}{c}{hard fusion (all)}    \\ \hline
  \multirow{2}{*}{T.1} & CV
  & 66.17 [61.7, 71.4]
  & 61.60 [56.7, 66.4]
  & 50.94 [46.1, 55.7]
  & 63.01 [59.6, 69.6]
  & 59.08 [54.5, 64.1]
  & 61.65 [56.8, 66.9]
  & 66.22 [62.8, 72.5]
  \\
  & Test
  & 54.89 [45.2, 63.4]
  & 46.05 [33.3, 55.7]
  & {\bf 59.18 [50.2, 68.7]}
  & 54.73 [46.1, 63.9]
  & 51.71 [39.4, 65.1]
  & 52.22 [42.6, 61.0]
  & 53.26  [44.7, 63.2]
  \\ \hline
  \multirow{2}{*}{T.2}
  & CV
  & 4.02 [3.6, 4.5]
  & 3.70 [3.3, 4.1]
  & 3.82 [3.5, 4.2]
  & 2.86 [2.5, 3.2]
  & 3.44 [3.1, 3.8]
  & 3.88 [3.5, 4.3]
  & 3.04, [2.7, 3.4]
  \\
  & Test
  & 3.82 [3.3, 4.3]
  & 4.48 [4.1, 4.9]
  & 3.76[3.2, 4.3]
  & \textbf{2.89 [2.3, 3.5]}
  & 3.73 [3.3, 4.2]
  & 3.45 [3.1, 3.8]
  & 3.08, [2.7, 3.5] \\  \hline
\end{tabular}}
\end{table*}

For the regression task, the comparable linguistic features on their
own proved to be the most effective features, with RMSE scores of 2.86 (r = 0.514)
and 2.89 (r = 0.337) for validation and test sets, respectively. Combining wav2vec
and linguistic features also proved effective, but the eGeMAPs
acoustic features were not found to be useful in this task.  Unlike
classification, regression results differed by language. For English,
the RMSE was 1.75, while for Chinese the RMSE was 3.71, reflecting the
standard deviations of MMSE (4.11 for Chinese and 1.27 for English). 



\section{Discussion}

The present dataset is considerably less heterogeneous in terms of
diagnoses and cognitive test scores than most public data used to
date in research on predictive models for cognitive function
assessment, including the few existing cross- and multi-lingual speech
datasets used in this area
\cite{bib:LuzHaiderEtAl23madress,bib:fraser2019multilingualMCI}. This
makes the learning tasks defined in this paper harder, as they need to
discriminate over a narrower range of values. However, 
our baseline models perform comparably to those models.

For the cognitive score prediction task (regression), we achieve an
RMSE score of 2.89, while \cite{bib:liu2021automatic}, for instance,
reports an RMSE of 3.76, but using only the English subset of our
data. The most comparable research is that conducted in a signal
processing grand challenge to generalise speech-based predictive
models across Greek and English
\cite{bib:LuzHaiderEtAl23madress,bib:LuzHaiderEtAl24admsp}. The best
performing models achieved a classification accuracy between 69\% and
87\% (AD vs NC) and a RMSE between 4.79 and 3.72 for MMSE score
prediction. However, these models involved training in one language
and testing in another, while our experimental setup yields comparable
results combining both languages in training and test, to our
knowledge, for the first time.

Our goal was not to push the state-of-the-art on these datasets, but
rather to establish proof-of-concept for our multilingual speech-only
model's capabilities to predict MMSE scores and detect MCI on a
homogeneous multilingual dataset. Given that (a) our baseline results
are comparable to other models in the literature (59.18\%UAR and 2.89
RMSE), (b) both MMSE prediction and MCI detection are relatively
uncommon compared to AD detection in the literature
\cite{bib:DelaFuenteRichieLuz2020JAD}, (c) this is the first
speech-only model built through combining datasets in two different
languages, and (d) the models seem to generalise well for diagnosis
across language (as suggested by the similarity in performance across the
classification tasks), we argue that our work significantly contributes
to the development of the field and will serve as a workable baseline
for the wider research community.

As a limitation of this study, it should be noted that MMSE has been
criticised for low discrimination (ceiling effect), especially in
preclinical dementia \cite{bib:carnero2014should}. This limitation is
also common in similar studies in this research field. Therefore,
future studies should aim to focus on other cognitive tests, more able
to discriminate early stages of cognitive impairment.

A distinctive characteristic of our approach is the use of
languages-agnostic and comparable languages-specific features. Our
results suggest that comparable linguistic features can be valuable in
MMSE prediction. While the fusion of wav2vec acoustic features to
linguistic features did not improve on the results obtained with
linguistic features alone, we believe that this approach should be
explored further in larger datasets.
\todo[inline,size=\normalsize]{ALL: here we will discuss our results,
  the limitations of the data and approaches, etc.

  Fasih: can you add some comments about the feature sets, and in
  particular about the notably superior performance of wav2vec over
  the alternatives in MMSE prediction?
}

\begin{table}[!htb]
\centering
\caption{Results Insights: Comparison of the best performing methods for the classification task (T.1) in \% UAR and the MMSE regression task (T.2) in RMSE across both languages }
\begin{tabular}{lllr}
  \hline
&  & T.1 & T.2  \\\hline
\multirow{2}{*}{English} &Cross-validation  &52.7  & 1.43 \\
                  & Test & 60.0 & 1.75  \\
\multirow{2}{*}{Chinese} & Cross-validation & 48.3 & 3.73  \\
  & Test & 60.4 & 3.71 \\
  \hline
\end{tabular}
\label{lan_results}
\end{table}


\section{Conclusion}

This paper presented a novel benchmark dataset for the development
and testing of models for cognitive assessment through automatic
analysis of connected speech. In particular, it defined learning tasks
for diagnosis of MCI and prediction of MMSE scores. A general processing
architecture for cross-lingual cognitive assessment was proposed which
encompassed language-agnostic acoustic features and comparable
linguistic features in a single predictive model for English and
Chinese speech. Baseline models illustrated these predictive tasks and
approach to feature extraction. The data and metadata have been made
available to the research community. With the increasing interest by
the medical community in speech biomarkers as a convenient and
cost-effective approach to early detection and monitoring of cognitive
problems, we expect this new resource will stimulate further research
in the little explored field of cross-lingual modelling of cognitive
function.
\mnSL{Please feel free add to and revise this section.}

\section{Acknowledgements}

\todo[inline,size=\normalsize]{Acknowledgement should only be included
  in the camera-ready version, not in the version submitted for
  review.  The 5th page is reserved exclusively for
  \red{acknowledgements} and references. No other content must appear
  on the 5th page.}

SL acknowledges support from the British Academy/Leverhulme
through Research Grant number SRG2223/231718. 

\bibliographystyle{IEEEtran}
\bibliography{taukadialpaper}

\begin{thebibliography}{10}
\providecommand{\url}[1]{#1}
\csname url@samestyle\endcsname
\providecommand{\newblock}{\relax}
\providecommand{\bibinfo}[2]{#2}
\providecommand{\BIBentrySTDinterwordspacing}{\spaceskip=0pt\relax}
\providecommand{\BIBentryALTinterwordstretchfactor}{4}
\providecommand{\BIBentryALTinterwordspacing}{\spaceskip=\fontdimen2\font plus
\BIBentryALTinterwordstretchfactor\fontdimen3\font minus
  \fontdimen4\font\relax}
\providecommand{\BIBforeignlanguage}[2]{{%
\expandafter\ifx\csname l@#1\endcsname\relax
\typeout{** WARNING: IEEEtran.bst: No hyphenation pattern has been}%
\typeout{** loaded for the language `#1'. Using the pattern for}%
\typeout{** the default language instead.}%
\else
\language=\csname l@#1\endcsname
\fi
#2}}
\providecommand{\BIBdecl}{\relax}
\BIBdecl

\bibitem{Nichols2022Feb}
{\relax GBD 2019 Dementia Forecasting Collaborators}, ``Estimation of the
  global prevalence of dementia in 2019 and forecasted prevalence in 2050,''
  \emph{Lancet Public Health}, vol.~7, no.~2, 2022.

\bibitem{bib:DelaFuenteRichieLuz2020JAD}
S.~de~la Fuente~Garcia, C.~Ritchie, and S.~Luz, ``Artificial intelligence,
  speech and language processing approaches to monitoring {Alzheimer's}
  disease: a systematic review,'' \emph{Journal of Alzheimer's Disease},
  vol.~78, no.~4, 2020.

\bibitem{bib:Nasrolahzadeh2018}
M.~Nasrolahzadeh, Z.~Mohammadpoory, and J.~Haddadnia, ``Higher-order spectral
  analysis of spontaneous speech signals in {Alzheimer's} disease,''
  \emph{Cognitive Neurodynamics}, vol.~12, no.~6, pp. 583--596, 2018.

\bibitem{bib:Mirheidari2019computational}
B.~Mirheidari, D.~Blackburn, R.~O'Malley, T.~Walker, A.~Venneri, M.~Reuber, and
  H.~Christensen, ``Computational cognitive assessment: Investigating the use
  of an intelligent virtual agent for the detection of early signs of
  dementia,'' in \emph{Procs. ICASSP}.\hskip 1em plus 0.5em minus 0.4em\relax
  IEEE, 2019, pp. 2732--2736.

\bibitem{bib:Mirzaei2018}
S.~Mirzaei, M.~{El Yacoubi}, S.~Garcia-Salicetti \emph{et~al.}, ``Two-stage
  feature selection of voice parameters for early {Alzheimer's} disease
  prediction,'' \emph{IRBM}, vol.~39, no.~6, pp. 430--435, 2018.

\bibitem{bib:Guo2019}
Z.~Guo, Z.~Ling, and Y.~Li, ``Detecting {Alzheimer's} disease from continuous
  speech using language models,'' \emph{Journal of Alzheimers Disease},
  vol.~70, no.~4, pp. 1163--1174, 2019.

\bibitem{bib:HaiderFuenteLuz20aspacf}
F.~Haider, S.~de~la Fuente, and S.~Luz, ``An assessment of paralinguistic
  acoustic features for detection of {Alzheimer's} dementia in spontaneous
  speech,'' \emph{IEEE J Sel Top Signal Process}, vol.~14, no.~2, pp. 272--281,
  2020.

\bibitem{bib:Prudhommeaux2015}
E.~T. Prud'hommeaux and B.~Roark, ``Graph-based word alignment for clinical
  language evaluation,'' \emph{Comput. Linguist.}, vol.~41, no.~4, pp.
  549--578, 2015.

\bibitem{bib:Sadeghian2017}
R.~Sadeghian, J.~D. Schaffer, and S.~A. Zahorian, ``{Speech Processing Approach
  for Diagnosing Dementia in an Early Stage},'' in \emph{Proc. Interspeech
  2017}, 2017, pp. 2705--2709.

\bibitem{bib:Shinkawa2019}
K.~Shinkawa, A.~Kosugi, M.~Nishimura, M.~Nemoto, K.~Nemoto, T.~Takeuchi,
  Y.~Numata, R.~Watanabe, E.~Tsukada, M.~Ota, S.~Higashi, T.~Arai, and
  Y.~Yamada, ``\BIBforeignlanguage{eng}{Multimodal behavior analysis towards
  detecting mild cognitive impairment: Preliminary results on gait and
  speech},'' \emph{\BIBforeignlanguage{eng}{Stud Health Technol Inform}}, vol.
  264, pp. 343--347, 2019.

\bibitem{jin2023consen}
L.~Jin, Y.~Oh, H.~Kim, H.~Jung, H.~J. Jon, J.~E. Shin, and E.~Y. Kim,
  ``{CONSEN}: Complementary and simultaneous ensemble for alzheimer’s disease
  detection and {MMSE} score prediction,'' in \emph{Proceedings of
  ICASSP}.\hskip 1em plus 0.5em minus 0.4em\relax IEEE, 2023, pp. 1--2.

\bibitem{bib:ostrand2021using}
R.~Ostrand and J.~Gunstad, ``Using automatic assessment of speech production to
  predict current and future cognitive function in older adults,''
  \emph{Journal of Geriatric Psychiatry and Neurology}, vol.~34, no.~5, pp.
  357--369, 2021.

\bibitem{bib:liu2021automatic}
Z.~Liu, L.~Proctor, P.~N. Collier, and X.~Zhao, ``Automatic diagnosis and
  prediction of cognitive decline associated with alzheimer’s dementia
  through spontaneous speech,'' in \emph{2021 IEEE International Conference on
  Signal and Image Processing Applications}.\hskip 1em plus 0.5em minus
  0.4em\relax IEEE, 2021, pp. 39--43.

\bibitem{bib:fu2020predicting}
Z.~Fu, F.~Haider, and S.~Luz, ``Predicting mini-mental status examination
  scores through paralinguistic acoustic features of spontaneous speech,'' in
  \emph{42nd Intl Conf of the IEEE Engineering in Medicine \& Biology Society},
  2020, pp. 5548--5552.

\bibitem{bib:Lopez-de-Ipina2015b}
K.~Lopez-de Ipi{\~{n}}a, J.~B. Alonso, J.~Sol{\'{e}}-Casals \emph{et~al.}, ``On
  automatic diagnosis of {Alzheimer's} disease based on spontaneous speech
  analysis and emotional temperature,'' \emph{Cognitive Computation}, vol.~7,
  no.~1, pp. 44--55, 2015.

\bibitem{bib:fraser2019multilingualMCI}
K.~C. Fraser, K.~{Lundholm Fors}, and D.~Kokkinakis, ``Multilingual word
  embeddings for the assessment of narrative speech in mild cognitive
  impairment,'' \emph{Computer Speech {\&} Language}, vol.~53, pp. 121--139,
  2019.

\bibitem{bib:fraser2019multilingualAD}
K.~C. Fraser, N.~Linz, B.~Li, K.~L. Fors, F.~Rudzicz, A.~K{\"o}nig,
  J.~Alexandersson, P.~Robert, and D.~Kokkinakis, ``Multilingual prediction of
  {Alzheimer's} disease through domain adaptation and concept-based language
  modelling,'' in \emph{Proceedings of NACL}, 2019, pp. 3659--3670.

\bibitem{bib:LuzHaiderEtAl23madress}
S.~Luz, F.~Haider, D.~Fromm, I.~Lazarou, I.~Kompatsiaris, and B.~MacWhinney,
  ``Multilingual {Alzheimer's} dementia recognition through spontaneous speech:
  a signal processing grand challenge,'' in \emph{Proccedings of ICASSP}.\hskip
  1em plus 0.5em minus 0.4em\relax IEEE Press, Jun. 2023.

\bibitem{albert2011diagnosis}
M.~S. Albert, S.~T. DeKosky, D.~Dickson, B.~Dubois, H.~H. Feldman, N.~C. Fox,
  A.~Gamst, D.~M. Holtzman, W.~J. Jagust, R.~C. Petersen \emph{et~al.}, ``The
  diagnosis of mild cognitive impairment due to {Alzheimer's} disease,''
  \emph{Alzheimer's \& dementia}, vol.~7, no.~3, pp. 270--279, 2011.

\bibitem{morris1993clinical}
J.~C. Morris, ``The clinical dementia rating {(CDR)} current version and
  scoring rules,'' \emph{Neurology}, vol.~43, no.~11, pp. 2412--2412, 1993.

\bibitem{bib:LanziSaylorEtAl23d}
A.~M. Lanzi, A.~K. Saylor, D.~Fromm, H.~Liu, B.~MacWhinney, and M.~L. Cohen,
  ``Dementiabank: Theoretical rationale, protocol, and illustrative analyses,''
  \emph{American Journal of Speech-Language Pathology}, vol.~32, no.~2, pp.
  426--438, 2023.

\bibitem{kaplan1983boston}
E.~Kaplan, \emph{Boston diagnostic aphasia examination booklet}.\hskip 1em plus
  0.5em minus 0.4em\relax Lea \& Febiger Philadelphia, PA, 1983.

\bibitem{nicholas1993system}
L.~E. Nicholas and R.~H. Brookshire, ``A system for quantifying the
  informativeness and efficiency of the connected speech of adults with
  aphasia,'' \emph{Journal of Speech, Language, and Hearing Research}, vol.~36,
  no.~2, pp. 338--350, 1993.

\bibitem{rockwell1947going}
N.~Rockwell, ``Going and coming [oil on canvas],'' \emph{Norman Rockwell Art
  Collection Trust, Indianapolis, IN, United States}, 1947.

\bibitem{nasreddine2005montreal}
Z.~S. Nasreddine, N.~A. Phillips, V.~B{\'e}dirian, S.~Charbonneau,
  V.~Whitehead, I.~Collin, J.~L. Cummings, and H.~Chertkow, ``The {Montreal}
  cognitive assessment, {MoCA}: a brief screening tool for mild cognitive
  impairment,'' \emph{Journal of the American Geriatrics Society}, vol.~53,
  no.~4, pp. 695--699, 2005.

\bibitem{bib:FasnachtWueestEtAl23cm}
J.~S. Fasnacht, A.~S. Wueest, M.~Berres, A.~E. Thomann, S.~Krumm, K.~Gutbrod,
  L.~A. Steiner, N.~Goettel, and A.~U. Monsch, ``Conversion between the
  montreal cognitive assessment and the mini-mental status examination,''
  \emph{Journal of the American Geriatrics Society}, vol.~71, no.~3, pp.
  869--879, 2023.

\bibitem{harris2019redcap}
P.~A. Harris, R.~Taylor, B.~L. Minor, V.~Elliott, M.~Fernandez, L.~O'Neal,
  L.~McLeod, G.~Delacqua, F.~Delacqua, J.~Kirby \emph{et~al.}, ``The {REDCap}
  consortium: building an international community of software platform
  partners,'' \emph{Journal of biomedical informatics}, vol.~95, p. 103208,
  2019.

\bibitem{bib:RosenbaumRubin83}
P.~R. Rosenbaum and D.~B. Rubin, ``{The central role of the propensity score in
  observational studies for causal effects},'' \emph{Biometrika}, vol.~70,
  no.~1, pp. 41--55, 04 1983.

\bibitem{bib:EybenSchererEtAl16gg}
F.~Eyben \emph{et~al.}, ``The {Geneva} minimalistic acoustic parameter set for
  voice research and affective computing,'' \emph{IEEE Trans Affect Computing},
  vol.~7, no.~2, 2016.

\bibitem{schneider2019wav2vec}
S.~Schneider, A.~Baevski, R.~Collobert, and M.~Auli, ``wav2vec: Unsupervised
  pre-training for speech recognition,'' \emph{arXiv preprint
  arXiv:1904.05862}, 2019.

\bibitem{bib:FerrerEieraCI24}
L.~Ferrer and P.~Riera, ``Confidence intervals for evaluation in machine
  learning,'' [Computer software], Accessed 1 March 2024,
  https://github.com/luferrer/ConfidenceIntervals.

\bibitem{bib:LuzHaiderEtAl24admsp}
S.~Luz, F.~Haider, D.~Fromm, I.~Lazarou, I.~Kompatsiaris, and B.~MacWhinney,
  ``An overview of the {ADReSS-M Signal Processing Grand Challenge on
  Multilingual Alzheimer's Dementia Recognition through Spontaneous Speech},''
  \emph{IEEE Open Journal of Signal Processing}, pp. 1--12, 2024.

\bibitem{bib:carnero2014should}
C.~Carnero-Pardo, ``Should the mini-mental state examination be retired?''
  \emph{Neurolog{\'\i}a (English Edition)}, vol.~29, no.~8, pp. 473--481, 2014.

\end{thebibliography}

\newpage

\end{document}